# Classification-based Financial Markets Prediction using Deep Neural Networks

Matthew Dixon[1], Diego Klabjan[2], and Jin Hoon Bang[3]

[1]Stuart School of Business, Illinois Institute of Technology, 10 West 35th Street, Chicago, IL 60616, matthew.dixon@stuart.iit.edu
[2]Department of Industrial Engineering and Management Sciences, Northwestern University, Evanston, IL, d-klabjan@northwestern.edu
[3]Department of Computer Science, Northwestern University, Evanston, IL, jinhoonbang@u.northwestern.edu

July 18, 2016

## Abstract

Deep neural networks (DNNs) are powerful types of artificial neural networks (ANNs) that use several hidden layers. They have recently gained considerable attention in the speech transcription and image recognition community (Krizhevsky et al., 2012) for their superior predictive properties including robustness to overfitting. However their application to algorithmic trading has not been previously researched, partly because of their computational complexity. This paper describes the application of DNNs to predicting financial market movement directions. In particular we describe the configuration and training approach and then demonstrate their application to backtesting a simple trading strategy over 43 different Commodity and FX future mid-prices at 5-minute intervals. All results in this paper are generated using a C++ implementation on the Intel Xeon Phi co-processor which is 11.4x faster than the serial version and a Python strategy backtesting environment both of which are available as open source code written by the authors.

## 1 Introduction

Many of the challenges facing methods of financial econometrics include non-stationarity, non-linearity or noisiness of the time series. While the application of artificial neural networks (ANNs) to time series methods are well documented (Faraway and Chatfield, 1998; Refenes, 1994; Trippi and DeSieno, 1992; Kaastra and Boyd, 1995) their proneness to over-fitting, convergence problems, and difficulty of implementation raised concerns. Moreover, their departure from the foundations of financial econometrics alienated the financial econometrics research community and finance practitioners.

However, algotrading firms employ computer scientists and mathematicians who are able to perceive ANNs as not merely black-boxes, but rather a non-parametric approach to modeling based on minimizing an entropy function. As such, there has been a recent resurgence in the method, in part facilitated by advances in modern computer architecture (Chen et al., 2013; Niaki and Hoseinzade, 2013; Vanstone and Hahn, 2010).

A deep neural network (DNN) is an artificial neural network with multiple hidden layers of units between the input and output layers. They have been popularized in the artificial intelligence community for their successful use in image classification (Krizhevsky et al., 2012) and speech recognition. The field is referred to as "Deep Learning".

In this paper, we shall use DNNs to partially address some of the historical deficiencies of ANNs. Specifically, we model complex non-linear relationships between the independent variables and dependent variable and reduced tendency to overfit. In order to do this we shall exploit advances in low cost many-core accelerator platform to train and tune the parameters of our model.

For financial forecasting, especially in multivariate forecasting analysis, the feed-forward topology has gained much more attention and shall be the approach used here. Back-propagation and gradient descent have been the preferred method for training these structures due to the ease of implementation and their tendency to converge to better local optima in comparison with other trained models. However, these methods can be computationally expensive, especially when used to train DNNs.

There are many training parameters to be considered with a DNN, such as the size (number of layers and number of units per layer), the learning rate and initial weights. Sweeping through the parameter space for optimal parameters is not feasible due to the cost in time and computational resources. We shall use mini-batching (computing the gradient on several training examples at once rather than individual examples) as one common approach to speeding up computation. We go further by expressing the back-propagation algorithm in a form that is amenable to fast performance on an Intel Xeon Phi co-processor (Jeffers and Reinders, 2013). General purpose hardware optimized implementations of the back-propagation algorithm are described by Shekhar and Amin (1994), however our approach is tailored for the Intel Xeon Phi co-processor.

The main contribution of this paper is to describe the application of deep neural networks to financial time series data in order to classify financial market movement directions. Traditionally, researchers will iteratively experiment with a handful of signals to train a level based method, such as vector autoregression, for each instrument (see for example Kaastra and Boyd (1995); Refenes (1994); Trippi and DeSieno (1992)). More recently, however, Leung et al. (2000) provide evidence that classification based methods outperform level based methods in the prediction of the direction of stock movement and trading returns maximization.

Using 5 minute interval prices from June 1989 to March 2013, our approach departs from the literature by using state-of-the-art parallel computing architecture to simultaneously train a single model from a large number of signals across multiple instruments, rather than using one model for each instrument. By aggregating the data across multiple instruments and signals, we enable the model to capture a richer set of information describing the time-varying co-movements across signals for each instrument price movement. Our results show that our model is able to predict the direction of instrument movement to, on average, 42% accuracy with a standard deviation across instruments of 11%. In some cases, we are able to predict as high as 68%. We further show how backtesting accuracy translates into the P&L for a simple long-only trading strategy and demonstrate sample mean Annualized Sharpe Ratios as high as 3.29 with a standard deviation of 1.12.

So in summary, our approach differs from other financial studies described in the literature in two distinct ways:

1. ANNs are applied to historical prices on an individual symbol and here 43 commodities and FX futures traded on the CME have been combined. Furthermore time series of lags, moving averages and moving correlations have been generated to capture memory and co-movements between symbols. Thus we have generated a richer dataset for the DNN to explore complex patterns.

2. ANNs are applied as a regression, whereas here the output is one of $\{-1, 0, 1\}$ representing a negative, flat or positive price movement respectively. The threshold for determining the zero state is set to $1 \times 10^{-3}$ (this is chosen to balance the class labels). The caveat is that restriction to a discrete set of output states may not replace a classical financial econometric technique, but it may be applicable for simple trading strategies which rely on the sign, and not the magnitude, of the forecasted price.

In the following section we introduce the back-propagation learning algorithm and use mini-batching to express the most compute intensive equations in matrix form. Once expressed in matrix form, hardware optimized numerical linear algebra routines are used to achieve an efficient mapping of the algorithm on to the Intel Xeon Phi co-processor. Section 3 describes the preparation of the data used to train the DNN. Section 4 describes the implementation of the DNN. Section 5 then presents results measuring the performance of a DNN. Finally in Section 6, we demonstrate the application of DNNs to backtesting using a walk forward methodology, and provide performance results for a simple buy-hold-sell strategy.

## 2 Deep Neural Network Classifiers

We begin with mathematical preliminaries. Let $\mathcal{D}$ denote the historical dataset of $M$ features and $N$ observations. We draw a training subset $\mathcal{D}_{\text{train}} \subset \mathcal{D}$ of $N_{\text{train}}$ observations and a test subset of $\mathcal{D}_{\text{test}} \subset \mathcal{D}$ of $N_{\text{test}}$ observations.

Denote the $n^{th}$ observation (feature vector) as $x_n \in \mathcal{D}_{\text{train}}$. In an ANN, each element of the vector becomes a node in the input layer, as illustrated in the figure below for the case when there are 7 input variables (features) per observation. In a fully connected feed-forward network, each node is connected to every node in the next layer. Although not shown in the figure, associated with each edge between the $i^{th}$ node in the previous layer and the $j^{th}$ node in the current layer $l$ is a weight $w_{ij}^{(l)}$.

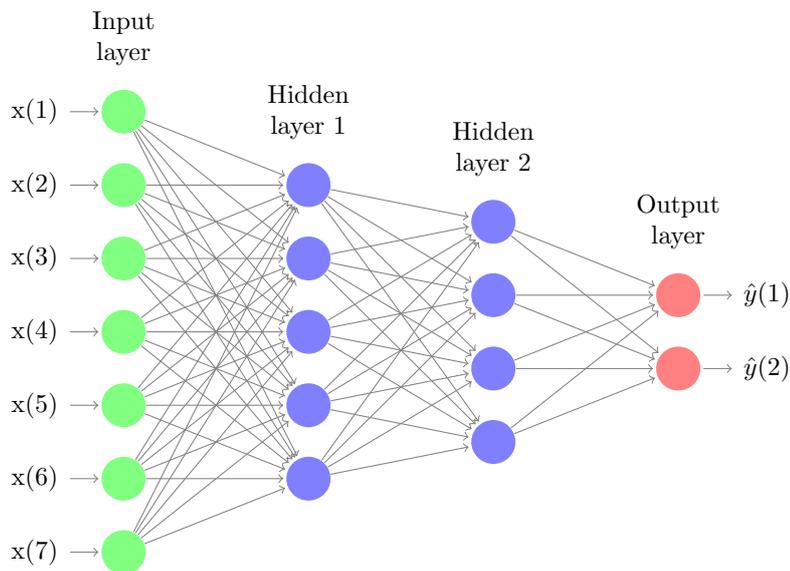

Figure 1: An illustrative example of a feed-forward neural network with two hidden layers, seven features and two output states. Deep learning network classifiers typically have many more layers, use a large number of features and several output states or classes. The goal of learning is to find the weight on every edge that minimizes the out-of-sample error measure.

In order to find optimal weightings $\mathbf{w} := \{\mathbf{w}^{(l)}\}_{l:=1 \to L}$ between nodes in a fully connected feed forward network with $L$ layers, we seek to minimize a cross-entropy function[1] of the form

$$E(\mathbf{w}) = -\sum_{n=1}^{N_{\text{test}}} e_n(\mathbf{w}), \qquad e_n(\mathbf{w}) := \sum_{k=1}^{K} y_{kn} ln\left(\hat{y}_{kn}\right). \qquad (1)$$

For clarity of exposition, we drop the subscript $n$. The binary target $y$ and output variables $\hat{y}$ have a 1-of-$k_s$ encoding for each symbol, where $y_k \in \{0, 1\}$ and $\sum_{k=i}^{i+k_s} \hat{y}_k = 1$ and $i = \{1, 1+k_s, 1+2k_s \ldots, K-k_s\}$, so that each output state associated with a symbol can be interpreted as a probabilistic weighting. To both ensure analytic gradient functions under the cross-entropy error measure and that the probabilities of each state sum to unity, the output layer is activated with a softmax function of the form

$$\hat{y}_k := \phi_{\text{softmax}}(\mathbf{s}^{(L)}) = \frac{\exp(s_k^{(L)})}{\sum_{j=1}^{k_s} \exp(s_j^{(L)})}, \qquad (2)$$

where for a fully connected feed-forward network, $s_j^{(l)}$ is the weighted sum of outputs from the previous layer $l-1$ that connect to node $j$ in layer $l$:

$$s_j^{(l)} = \sum_{i=1}^{n_{(l-1)}} w_{ij}^{(l)} x_i^{(l-1)} + \text{bias}_j^{(l)}, \qquad (3)$$

where $n_{(l)}$ are the number of nodes in layer $l$. The gradient of the likelihood function w.r.t. $s_k^{(L)}$ takes the simple form:

$$\frac{\partial e(\mathbf{w})}{\partial s_k^{(L)}} = \hat{y}_k - y_k. \qquad (4)$$

---

[1] The use of entropy in econometrics research has been well established (see for example Golan et al. (1996); Wu and Perloff (2007)).

The recursion relation for the back propagation using conjugate gradients is:

$$\delta_i^{(l-1)} = \sum_{j=1}^{n_{(l-1)}} \delta_j^{(l)} w_{ij}^{(l)} \sigma(s_i^{(l-1)})(1 - \sigma(s_i^{(l-1)})), \qquad (5)$$

where we have used the analytic form of the derivative of the sigmoid function

$$\sigma'(v) = \sigma(v)(1 - \sigma(v)) \qquad (6)$$

to activate all hidden layer nodes. So in summary, a trained feed-forward network can be used to predict the probability of an output state (or class) for each of the symbols concurrently, given any observation as an input, by recursively applying Equation 3. The description of how the network is trained now follows.

**Stochastic Gradient Descent** Following Rojas (1996), we now revisit the backpropagation learning algorithm based on the method of stochastic gradient descent (SGD) algorithm. Despite only being first order, SGD serves as the optimization method of choice for DNNs due to the highly non-convex form of the utility function (see for example Li et al. (2014)). After random sampling of an observation $i$, the SGD algorithm updates the parameter vector $\mathbf{w}^{(l)}$ for the $l^{th}$ layer using

$$\mathbf{w}^{(l)} = \mathbf{w}^{(l)} - \gamma \nabla E_i(\mathbf{w}^{(l)}), \qquad (7)$$

where $\gamma$ is the learning rate. A high level description of the sequential version of the SGD algorithm is given in Algorithm 1. Note that for reasons of keeping the description simple, we have avoided some subtleties of the implementation.

**Algorithm 1** STOCHASTIC GRADIENT DESCENT

1: $\mathbf{w} \leftarrow \mathbf{r}$, $r_i \in \mathcal{N}(\mu, \sigma)$, $\forall i$
2: $E \leftarrow 0$
3: **for** $i = 0$ to $n - 1$ **do**
4: $\quad E \leftarrow E + E_i(\mathbf{w})$
5: **end for**
6: **while** $E \geq \tau$ **do**
7: $\quad$ **for** $t = 0$ to $n - 1$ **do**
8: $\quad\quad i \leftarrow$ sample with replacement in $[0, n-1]$
9: $\quad\quad \mathbf{w} \leftarrow \mathbf{w} - \gamma \nabla E_i(\mathbf{w})$
10: $\quad$ **end for**
11: $\quad E \leftarrow 0$
12: $\quad$ **for** $i = 0$ to $n - 1$ **do**
13: $\quad\quad E \leftarrow E + E_i(\mathbf{w})$
14: $\quad$ **end for**
15: **end while**

## 2.1 Mini-batching

It is well known that mini-batching improves the computational performance of the feedforward and back-propagation computations (Shekhar and Amin, 1994; Li et al., 2014). We process $b$ observations in one mini-batch. This results in a change to the SGD algorithm and the dimensions of data-structures that are used to store variables. In particular, $\delta$, $x$, $s$ and $E$ now have a batch dimension. Note however that the dimensions of $w^{(l)}$ remain the same. The above equations can be now be modified.

With slight abuse of notation, we redefine the dimension $\delta^{(l)}, X^{(l)}, S^{(l)} \in \mathbf{R}^{n_l \times b}$, $\forall l$, $E \in \mathbf{R}^{n_{(L)} \times b}$, where $b$ is the size of the mini-batch.

Crucially for computational performance of the mini-batching, the computation of the sum in each layer of the feed-forward network can be expressed as a matrix-matrix product:

$$S^{(l)} = \left(X_i^{(l-1)}\right)^T \mathbf{w}^{(l)}. \tag{8}$$

For the $i^{th}$ neuron in output layer $L$ and the $j^{th}$ observation in the mini-batch

$$\delta_{ij}^{(L)} = \sigma_{ij}^{(L)}(1 - \sigma_{ij}^{(L)}) E_{ij}. \tag{9}$$

For all intermediate layers $l < L$, the recursion relation for $\delta$ is

$$\delta_{ij}^{(l-1)} = \sigma_{ij}^{(l)}(1 - \sigma_{ij}^{(l)}) w_{ij}^{(l)} \delta_{ij}^{(l)}. \tag{10}$$

The weights are updated with matrix-matrix products for each layer

$$\Delta \mathbf{w}^{(l)} = \gamma X^{(l-1)} \left(\delta^{(l)}\right)^T. \tag{11}$$

## 3 The Data

Our historical dataset contains 5 minute mid-prices for 43 CME listed commodity and FX futures from March 31st 1991 to September 30th, 2014. We use the most recent fifteen years of data because the previous period is less liquid for some of the symbols, resulting in long sections of 5 minute candles with no price movement. Each feature is normalized by subtracting the mean and dividing by the standard deviation. The training set consists of 25,000 consecutive observations and the test set consists of the next 12,500 observations. As described in Section 6, these sets are rolled forward ten times from the start of the liquid observation period, in 1000 observation period increments, until the final 37,500 observations from March 31st, 2005 until the end of the dataset.

The overall training dataset consists of the aggregate of feature training sets for each of the symbols. The training set of each symbol consists of price differences and engineered features including lagged prices differences from 1 to 100, moving price averages with window sizes from 5 to 100, and pair-wise correlations between the returns and the returns of all other symbols. The overall training set contains 9895 features. The motivation for including these features in the model is to capture memory in the historical data and co-movements between symbols.

## 4 Implementation

The architecture of our network contains five learned fully connected layers. The first of the four hidden layers contains 1000 neurons and each subsequent layer is tapered by 100. The final layer contains 129 output neurons - three values per symbol of each of the 43 futures contracts. The result of including a large number of features and multiple hidden layers is that there are 12,174,500 weights in total.

The weights are initialized with an Intel MKL VSL random number generator implementation that uses the Mersenne Twistor (MT19937) routine. Gaussian random numbers are generated from transforming the uniform random numbers with an inverse Gaussian cumulative distribution function with zero mean and standard deviation of 0.01. We initialized the neuron biases in the hidden layers with the constant 1.

We used the same learning rate for all layers. The learning rate was adjusted according to a heuristic which is described in Algorithm 2 below and is similar to the approach taken by Krizhevsky et al. (2012) except that we use cross entropy rather than the validation error. We sweep the parameter space of the learning rate from $[0.1, 1]$ with increments of 0.1. We further divide the learning rate $\gamma$ by 2 if the cross-entropy does not decrease between epochs.

In Algorithm 2, the subset of the training set used for each epoch is defined as
$$\mathcal{D}_e := \{x_{n_k} \in \mathcal{D}_{\text{train}} \mid n_k \in \mathcal{U}(1, N_{\text{train}}), k := 1, \ldots, N_{\text{epoch}}\} \qquad (12)$$

**Algorithm 2** DEEP LEARNING METHODOLOGY
---
1: **for** $\gamma := 0.1, 0.2, \ldots, 1$ **do**
2: $\quad w_{i,j}^{(l)} \leftarrow r,\ r \in \mathcal{N}(\mu, \sigma),\ \forall i, j, l$ ▷ Initialize all weights
3: $\quad$ **for** $e = 1, \ldots, N_e$ **do** ▷ Iterate over epochs
4: $\quad\quad$ Generate $\mathcal{D}_e$
5: $\quad\quad$ **for** $m = 1, \ldots, M$ **do** ▷ Iterate over mini-batches
6: $\quad\quad\quad$ Generate $\mathcal{D}_m$
7: $\quad\quad\quad$ **for** $l = 2, \ldots, L$ **do**
8: $\quad\quad\quad\quad$ Compute all $x_j^{(l)}$ ▷ Feed-Forward network construction
9: $\quad\quad\quad$ **end for**
10: $\quad\quad\quad$ **for** $l = L, \ldots, 2$ **do**
11: $\quad\quad\quad\quad$ Compute all $\delta_j^{(l)} := \nabla_{s_j^{(l)}} E$ ▷ Backpropagation
12: $\quad\quad\quad\quad$ Update the weights: $\mathbf{w}^{(l)} \leftarrow \mathbf{w}^{(l)} - \gamma X^{(l-1)} \left(\delta^{(l)}\right)^T$
13: $\quad\quad\quad$ **end for**
14: $\quad\quad$ **end for**
15: $\quad$ **end for**
16: $\quad$ If cross_entropy(e) $\leq$ cross_entropy(e-1) then $\gamma \leftarrow \gamma/2$
17: **end for**
18: Return final weights $w_{i,j}^{(l)}$
---

and the mini-batch with in each epoch set is defined as

$$\mathcal{D}_m := \{x_{n_k} \in \mathcal{D}_{ep} \mid n_k \in \mathcal{U}(1, N_{\text{epoch}}), k := 1, \ldots, N_{\text{mini-batch}}\}. \quad (13)$$

As mentioned earlier, the mini-batching formulation of the algorithm facilitates efficient parallel implementation, the details and timings of which are described by Dixon et al. (2015). The overall time to train a DNN on an Intel Xeon Phi using the data described above is approximately 8 hours when factoring in time for calculation of error measures on the test set and thus the training can be run as an overnight batch job should daily retraining be necessary. This is 11.4x faster than running the serial version of the algorithm.

## 5 Results

This section describes the backtesting of DNNs for a simple algo-trading strategy. The purpose is to tie together classification accuracy with strategy performance measurements and is not intended to provided an exhaustive exploration of trading strategies or their performance. For each symbol, we calculate the classification accuracies for each 130 day moving test window. This is repeated to give a set of ten classification errors. Figure 2 shows a box plot of the classification accuracy of the DNN for all the 43 CME Commodity and FX futures. Each symbol is represented by a box and whisker vertical bar - the box represents the region between the lower and upper quartiles of the sample distribution of classification accuracies. The median of the sample distribution of classification accuracies is represented as a red horizontal line.

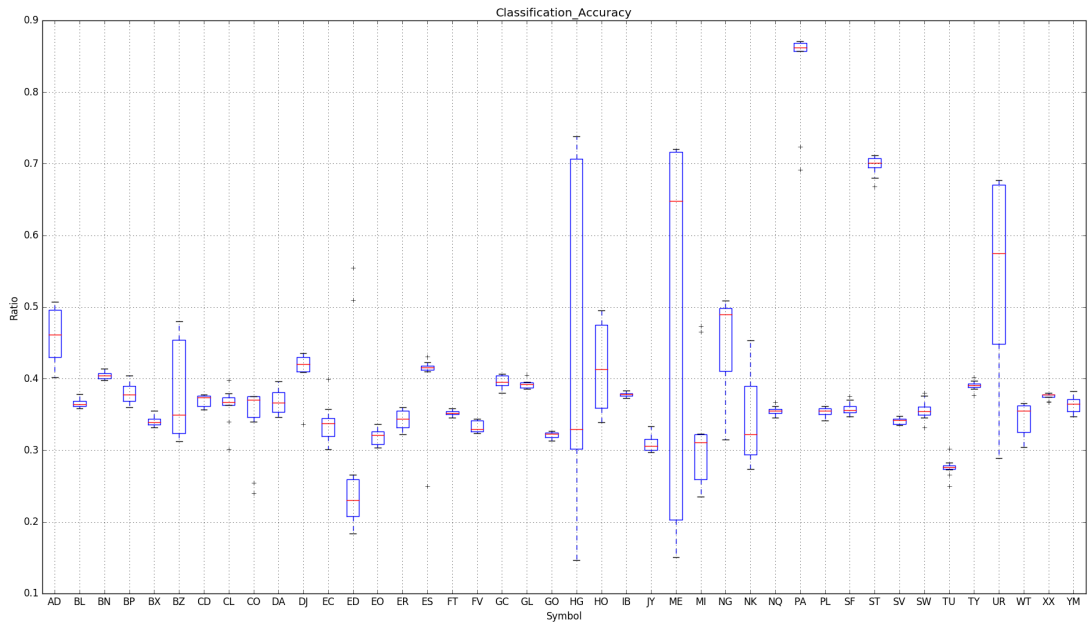

Figure 2: This figure shows the classification accuracy of the DNN applied to 43 CME Commodity and FX futures. Each symbol is represented by a box and whisker vertical bar - the box represents the region between the lower and upper quartiles of the sample distribution of classification accuracies. The median of the sample distribution of classification accuracies is represented as a red horizontal line.

Figure 3 below shows the distribution of the average classification accuracy over 10 samples of the DNN across the 43 CME Commodity and FX futures. There's a heavier density around an accuracy of 0.35 which is slightly better than a random selection.

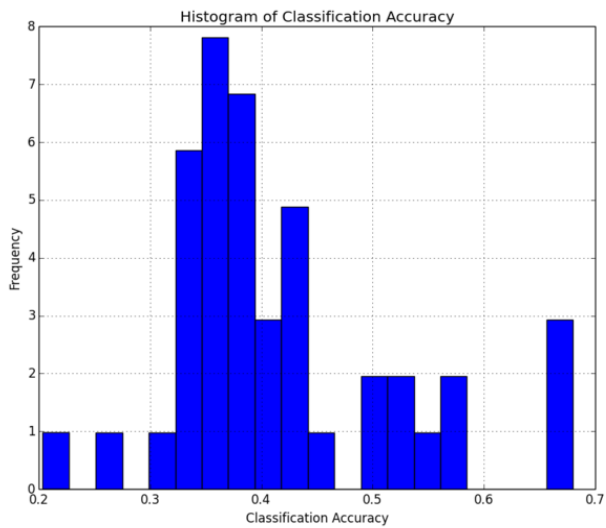

Figure 3: This figure shows the distribution of the average classification accuracy of the DNN applied to 43 CME Commodity and FX futures.

Table 1 shows the top five instruments for which the sample mean of the classification rate was highest on average over the ten walk forward experiments. Also shown are the F1-scores ('harmonic means') which are considered to be a more robust measure of performance due to less sensitivity to class imbalance than classification accuracies. The mean and standard deviation of the sample averaged classification accuracies and F1-scores over the 43 futures are also provided.

| Symbol | Futures | Classification Accuracy | F1-score |
|---|---|---|---|
| HG | Copper | 0.68 | 0.59 |
| ST | Transco Zone 6 Natural Gas (Platts Gas Daily) Swing | 0.67 | 0.54 |
| ME | Gulf Coast Jet (Platts) Up-Down | 0.67 | 0.54 |
| TU | Gasoil 0.1 Cargoes CIF NWE (Platts) vs. Low Sulphur Gasoil | 0.56 | 0.52 |
| MI | Michigan Hub 5 MW Off-Peak Calendar-Month Day-Ahead Swap | 0.55 | 0.5 |
| mean | - | 0.42 | 0.37 |
| std | - | 0.11 | 0.1 |

Table 1: This table shows the top five instruments for which the sample mean of the classification rate was highest over the ten walk forward experiments. F1-scores are also provided. The mean and standard deviation of the sample mean classification accuracies and F1-scores over the 43 futures are also provided.

Note that the worst five performing instruments performed no better or even worse than white noise on average over the ten experiments.

## 6 Strategy Backtesting

The paper has thus far considered the predictive properties of the deep neural network. Using commodity futures historical data at 5 minute intervals over the period from March 31st 1991 to September 30th, 2014, this section describes the application of a walk forward optimization approach for backtesting a simple trading strategy.

Following the walk forward optimization approach described in Tomasini and Jaekle (2011), an initial optimization window of 25,000 5-minute observation periods or approximately 260 days (slightly more than a year) is chosen for training the model using all the symbol data and their engineered time series. The learning rate range is swept to find the model which gives the best out-of-sample prediction rate - the highest classification rate on the out-of-sample ('hold-out') set consisting of 12,500 consecutive and more recent observations.

Using the optimized model, the expected P&L of the trading strategy is then evaluated over the out-of-sample period consisting of 12,500 consecutive 5-minute observation periods or approximately 130 days. Even though all symbols are trained together using one DNN model, the cumulative P&L is calculated independently for each symbol. As illustrated in Figure 4, this step is repeated by sliding the training window forward by 1000 observation periods and repeating the out-of-sample error analysis and strategy performance measurement for ten windows.

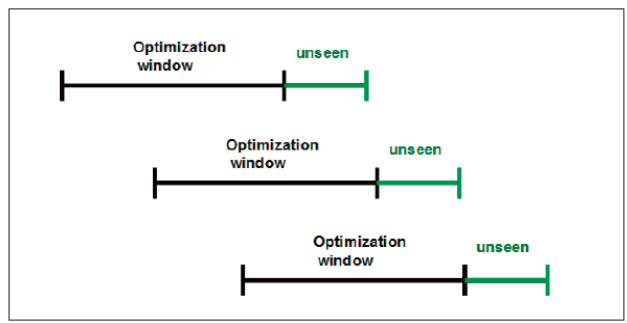

Figure 4: An illustration of the walk forward optimization method used for backtesting the strategy
.

## 6.1 Example trading strategy

In order to demonstrate the application of DNNs to algorithmic trading, a simple buy-hold-sell intraday trading strategy is chosen contingent on whether the instrument price is likely to increase, be neutral, or decrease over the next time interval respectively. For simplicity, the strategy only places one lot market orders. The strategy closes out a short position and takes a long position if the label is 1, holds the position if the label is zero and closes out the long position and takes a short position if the label is -1. In calculating the cumulative unrealized P&L, the following simplifying assumptions are made:

- the account is opened with $100k of USD;

- there is sufficient surplus cash available in order to always maintain the brokerage account margin, through realization of the profit or otherwise;

- there are no limits on the minimum or maximum holding period and positions can be held overnight;

- the margin account is assumed to accrue zero interest;

- transaction costs are ignored;

- no operational risk measures are deployed, such as placing stop-loss orders.

- the market is always sufficiently liquid that a market order gets filled immediately at the mid-price listed at 5 minute intervals and so slippage effects are ignored; and

- The placing of 1 lot market orders at 5 minute intervals has no significant impact on the market and thus the forecast does not account for limit order book dynamics in response to the trade execution.

These assumptions, especially those concerning trade execution and absence of live simulation in the backtesting environment are of course inadequate to demonstrate alpha generation capabilities of the DNN based strategy but serve as a starting point for commercial application of this research.

Returns of the strategy are calculated by first aggregating intraday P&L changes to daily returns and then annualizing them. Figure 5 show a box plot of the sample distribution of the time-averaged daily returns of the strategy applied separately to each of the 43 CME front month Commodity and FX futures over the 130 day trading horizons

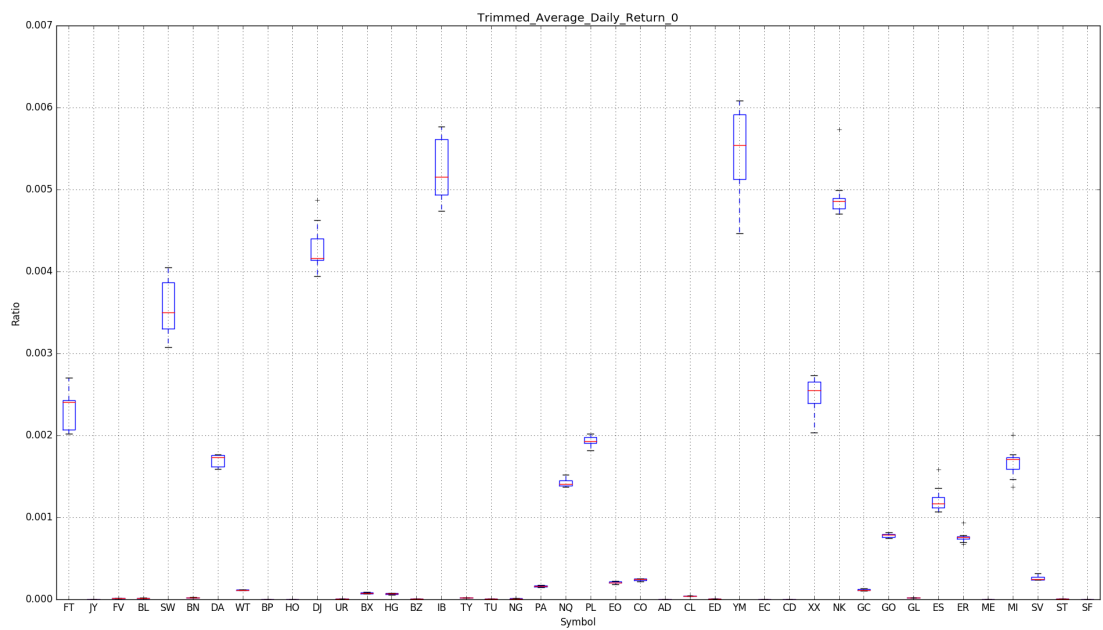

Figure 5: This figure shows a box plot of the sample distribution of the time-averaged daily returns of the strategy applied separately to each of the 43 CME Commodity and FX futures over the 130 day trading horizons. The red square with a black border denotes the sample average for each symbol.

Figure 6 compares the cumulative unrealized net dollar profit of the strategy for the case when perfect forecasting information is available ('perfect foresight') against using the DNN prediction ('predict'). The graph is shown for one 130 day trading horizon for front month Platinum (PL) futures.

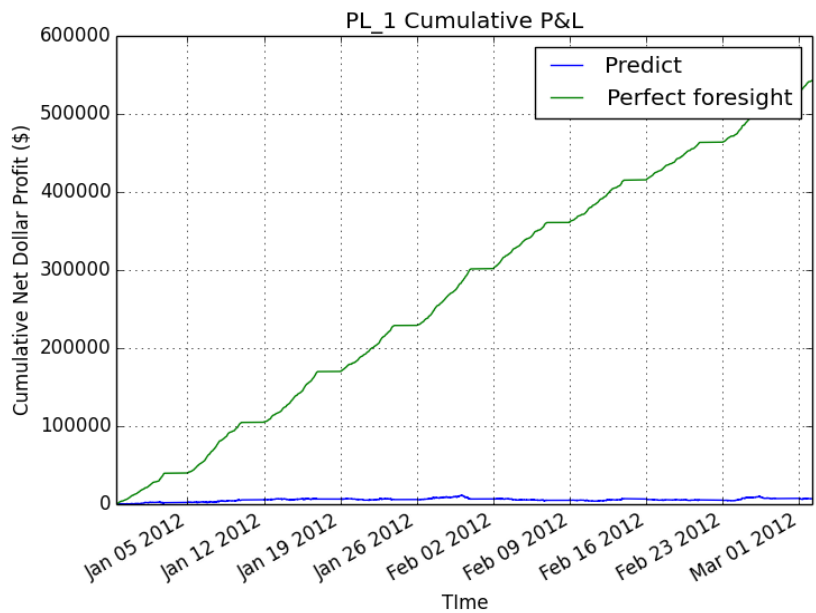

Figure 6: This figure shows the cumulative unrealized net dollar profit of a simple strategy. In order to quantify the impact of information loss, the profit under perfect forecasting information is denoted as 'perfect foresight' (green line) and the profit using the DNN prediction is denoted as 'predict' (blue line). The graph is shown for one 130 day trading horizon in front month Platinum (PL) futures.

Figure 7 shows a box plot of the maximum drawdown of a simple strategy applied over ten walk forward experiments for each symbol.

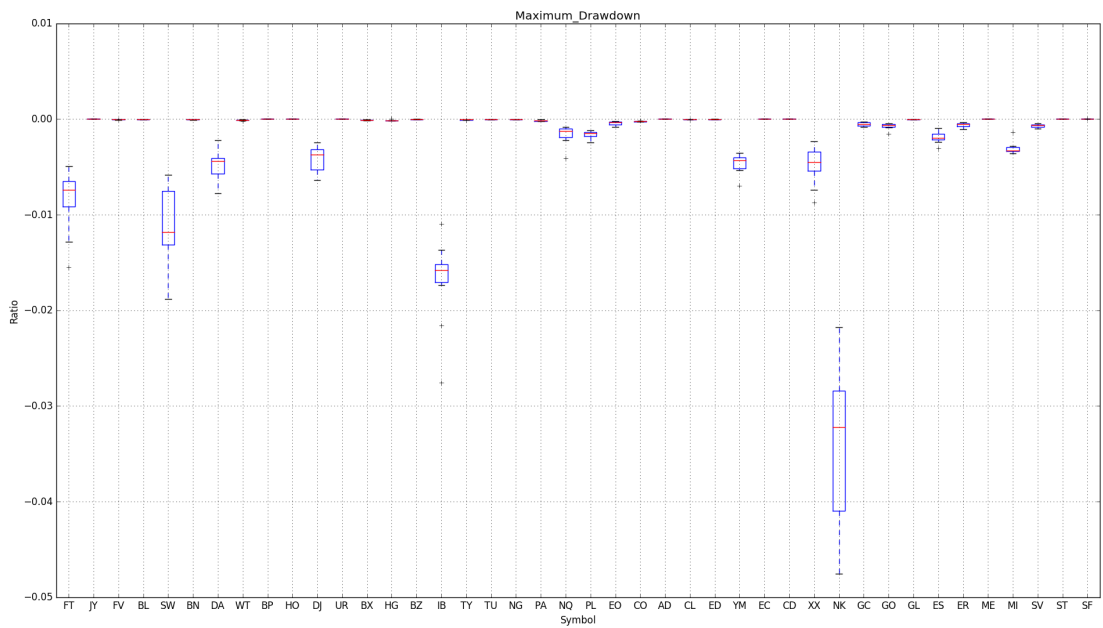

Figure 7: This figure shows a box plot of the maximum drawdown of a simple strategy applied over ten walk forward experiments for each symbol.

Figure 8 shows the range of annualized Sharpe ratios measured over each moving period of 12,500 observation points for the top five performing futures contracts[2]. This figure is also supplemented by Table 2 which shows the top five instruments for which the sample mean of the annualized Sharpe ratio was highest over the ten walk forward experiments. The values in parentheses denote the standard deviation over the ten experiments. Also shown, are the sample mean and standard deviations of the Capability ratios (where $n = 130$) under the assumption of normality of returns.

---

[2]No benchmark has been used in the calculation of the Sharpe ratios.

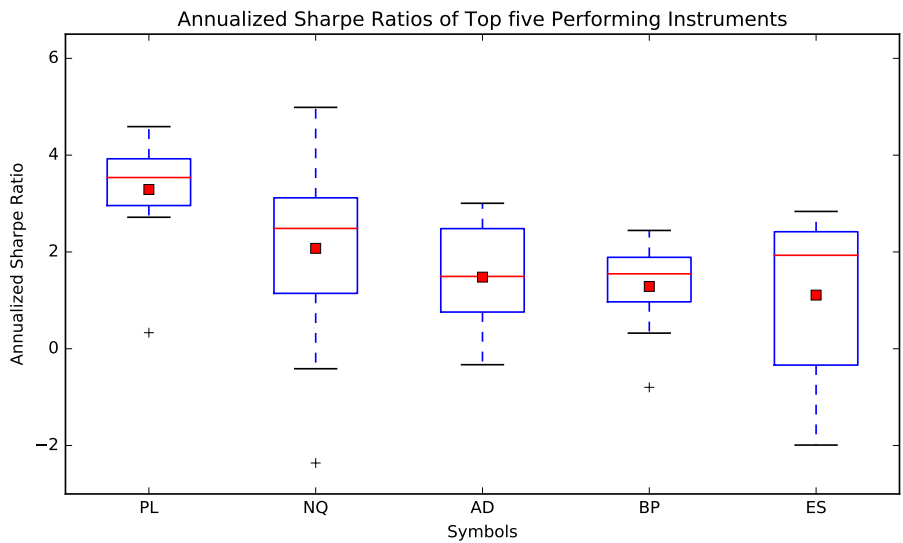

Figure 8: This figure shows a box plot of the distribution of the annualized Sharpe ratios sampled over ten walk forward experiments of 12,500 observation points. Only the top five performing futures contracts have been considered. The simple trading strategy is described above. Key: PL: Platinum, NQ: E-mini NASDAQ 100 Futures, AD: Australian Dollar, BP: British Pound, ES: E-mini S&P 500 Futures.

| Symbol | Futures | Annualized Sharpe Ratio | Capability Ratio |
|--------|---------|-------------------------|------------------|
| PL | Platinum | 3.29 (1.12 ) | 12.51 (4.27) |
| NQ | E-mini NASDAQ 100 Futures | 2.07 (2.11) | 7.89 (8.03) |
| AD | Australian Dollar | 1.48 (1.09) | 5.63 (4.13) |
| BP | British Pound | 1.29 (0.90) | 4.90 (3.44) |
| ES | E-mini S&P 500 Futures | 1.11 (1.69 ) | 4.22 (6.42) |

Table 2: This table shows the top five instruments for which the mean annualized Shape ratio was highest on average over the ten walk forward optimizations. The values in parentheses denote the standard deviation over the ten experiments. Also shown, are the mean and standard deviation of the Capability ratios under the assumption of normality of returns.

Table 3 lists the initial margin, maintenance margin and contract size specified by the CME used to calculate the cumulative unrealized P&L and strategy performance for the top five performing futures positions.

| Symbol | initial margin | maint. margin | contract size |
|--------|----------------|---------------|---------------|
| PL | 2090 | 1900 | 50 |
| NQ | 5280 | 4800 | 50 |
| AD | 1980 | 1800 | 100000 |
| BP | 2035 | 1850 | 62500 |
| ES | 5225 | 4750 | 50 |

Table 3: This table lists the initial margin, maintenance margin and contract size specified by the CME used to calculate the cumulative P&L and strategy performance for the top five performing futures positions.

Table 4 shows the correlation of the daily returns of the strategy on each of the five most liquid instruments in the list of 43 CME futures with their relevant ETF benchmarks. The values represent the summary statistics of the correlations over the ten experiments. When averaged over ten experiments, the strategy returns are observed to be weakly correlated with the benchmark returns and, in any given experiment, the absolute value of the correlations are all under 0.5.

| Symbol | Benchmark ETF | Correlation | | | |
|--------|---------------|------|-----------|-----|-----|
| | | Mean | Std. Dev. | Max | Min |
| NQ | PowerShares QQQ ETF (QQQ) | 0.013 | 0.167 | 0.237 | -0.282 |
| DJ | SPDR Dow Jones Industrial Average ETF (DIA) | 0.008 | 0.194 | 0.444 | -0.257 |
| ES | SPDR S&P 500 ETF (SPY) | -0.111 | 0.110 | 0.057 | -0.269 |
| YM | SPDR Dow Jones Industrial Average ETF (DIA) | -0.141 | 0.146 | 0.142 | -0.428 |
| EC | CurrencyShares Euro ETF (FXE) | -0.135 | 0.108 | 0.154 | -0.229 |

Table 4: This table shows the correlation of the daily returns of the strategy on each of the five most liquid instruments in the list of 43 CME futures with their relevant ETF benchmarks. The values represent the summary statistics of the correlations over the ten experiments. Key: NQ: E-mini NASDAQ 100 Futures, DJ: DJIA ($10) Futures, ES: E-mini S&P 500 Futures, YM: E-mini Dow ($5) Futures, EC: Euro FX Futures.

# 7 Conclusion

Deep neural networks (DNNs) are a powerful type of artificial neural network (ANN) that use several hidden layers. In this paper we describe the implemen-tation and training of DNNs. We observe, for a historical dataset of 5 minute mid-prices of multiple CME listed futures prices and other lags and filters that DNNs have substantial predictive capabilities as classifiers if trained concurrently across several markets on labelled data. We further demonstrate the application of DNNs to backtesting a simple trading strategy and demonstrate the prediction accuracy and its relation to the strategy profitability. All results in this paper are generated using a C++ implementation on the Intel Xeon Phi co-processor which is 11.4x faster than the serial version and a Python strategy backtesting environment both of which are available as open source code written by the authors.

# 8 Acknowledgements

The authors gratefully acknowledge the support of Intel Corporation in funding this research and the anonymous reviewers for useful comments.